\documentclass{article} 
\usepackage{iclr2027/iclr2027_conference,times}
\usepackage{subfiles}

\usepackage{amsmath,amsfonts,bm}

\def\eqref#1{equation~\ref{#1}}

\def\1{\bm{1}}

\DeclareMathAlphabet{\mathsfit}{\encodingdefault}{\sfdefault}{m}{sl}
\SetMathAlphabet{\mathsfit}{bold}{\encodingdefault}{\sfdefault}{bx}{n}

\usepackage{hyperref}
\usepackage{url}
\usepackage{tabularx}

\title{MemPrice: Online Memory Allocation via Adaptive Pricing for Long-Horizon Agents}

\author{
Yiming Zong\thanks{Equal contribution.}
\quad
Shiyuan Piao\footnotemark[1]
\quad
Jiashuo Jiang\thanks{Corresponding author.}
\quad
Fugee Tsung\footnotemark[2]
\\
Department of Industrial Engineering \& Decision Analysis, HKUST
\\
\texttt{\{yzongac,shiyuan.piao\}@connect.ust.hk}
\quad
\texttt{\{jsjiang,season\}@ust.hk}
}

\usepackage{algorithm}
\usepackage{algorithmic}
\usepackage{cleveref}
\usepackage{booktabs}
\usepackage{multirow}
\usepackage{makecell}
\usepackage[table]{xcolor}
\definecolor{oursbg}{HTML}{F1EFFE}

\usepackage{graphicx}
\usepackage{wrapfig}
\usepackage{comment}

\usepackage{amsmath,amssymb,amsthm,mathtools}
\theoremstyle{plain}
\newtheorem{theorem}{Theorem}
\newtheorem{proposition}{Proposition}

\theoremstyle{definition}

\newtheorem{assumption}{Assumption}

\theoremstyle{remark}

\iclrfinalcopy 
\begin{document}

\maketitle
\lhead{Preprint}
\begin{abstract}
Long-horizon LLM agents face a fundamental allocation problem as interaction histories grow: which information to retain, how much context to devote to it, and at what granularity. Existing memory mechanisms largely focus on relevance, retrieval, or compression, leaving this allocation trade-off implicit. We introduce \textsc{MemPrice}, which treats active context as a shared resource and dynamically allocates it across memories and representation levels. \textsc{MemPrice} turns accumulated memory pressure into an adaptive price on context, encouraging detail only when its estimated retention value justifies the additional footprint. This pressure persists across memory refreshes, allowing the contents of memory to evolve while preserving long-run control over context usage.
We show that irreversible compression can force linear worst-case regret against hindsight following its own memory trajectory, and introduce a same-state benchmark to evaluate allocation on \textsc{MemPrice}'s realized memory states. \textsc{MemPrice} enforces the hard context budget at every round and regulates long-run occupancy through adaptive pricing, without prior knowledge of the interaction horizon. Under priced-frontier regularity, its average allocation-regret bound against this benchmark vanishes when accumulated estimation error is sublinear.
We evaluate the framework under matched memory budgets on LongMemEval, LoCoMo, AppWorld, and WebShop across three foundation models, spanning both query-blind memory construction and goal-directed agent context management.
The complete experimental code and reproduction scripts are available in our anonymous repository at \url{https://anonymous.4open.science/r/ICLR2027-MemPrice}.
\end{abstract}

\section{Introduction}
\label{sec:introduction}
Foundation models are increasingly deployed as long-horizon agents that reason, act, and interact over extended trajectories \citep{yao2023react,shinn2023reflexion}. As these interactions grow, the accumulated history can quickly exceed the active context available to the model, making memory management a central component of agent design. Existing systems address this challenge through persistent memory, hierarchical storage, retrieval, summarization, and adaptive context construction \citep{park2023generative,packer2023memgpt,xu2025amem}. More recent methods further predict which past information will matter for upcoming decisions or dynamically compress interaction trajectories to control context growth \citep{wei2026pace,ye2026agentfold}. Together, these advances substantially improve an agent's ability to operate beyond a fixed context window.

However, most existing approaches formulate memory management primarily as a problem of \emph{retrieval, relevance estimation, or compression}, leaving the underlying allocation of context capacity largely implicit. 
Knowing that a memory is relevant does not determine whether it is worth the space it consumes: a detailed record may provide only slightly more value than a compact summary while occupying substantially more context, and retaining one item necessarily reduces the capacity available to others. Similarity, importance, or predicted relevance scores therefore do not directly capture the opportunity cost of memory occupancy. Moreover, this trade-off changes over time as new information arrives and the user's informational needs evolve. Once information has been compressed or discarded, early allocation mistakes may also be irreversible until the memory is refreshed. We argue that these effects are better understood by treating active context explicitly as a limited resource: rather than only asking \emph{what information is relevant}, an agent must decide \emph{which information deserves memory, at what granularity, and at what cost}.

To this end, we introduce \textsc{MemPrice}, a memory-allocation framework that makes the opportunity cost of active context explicit. Each memory can be retained at multiple levels of detail, while a persistent virtual queue turns accumulated occupancy pressure into a shared price on context. \textsc{MemPrice} then jointly chooses representations under the hard context budget, retaining finer detail only when its estimated retention value warrants the additional footprint. Crucially, the queue persists across memory refreshes, so retrieval can change what is brought back into context without erasing the resource pressure accumulated over the interaction. 
Irreversible compression changes which representations remain available, so hindsight may benefit from having preserved different information, not just from better allocation. We therefore evaluate allocation on \textsc{MemPrice}'s realized memory-state trajectory, while still allowing hindsight to redistribute occupancy across time. \textsc{MemPrice} always satisfies hard capacity and, under bounded utilities and estimation error, drives excess average occupancy to zero without knowing the interaction horizon. Under priced-frontier regularity, its average-regret bound against this same-state benchmark also vanishes when estimation error accumulates sublinearly.
We evaluate the framework under two complementary regimes: \emph{Query-Blind Memory}, where the eventual question is revealed only after a long interaction, and \emph{Goal-Observable Memory}, where the target is known from the outset and relevant evidence must be accumulated over time.
Our main contributions are:
\begin{itemize}
\item We formulate long-horizon agent memory management as an explicit online resource-allocation problem, and propose \textsc{MemPrice}, which combines capacity-feasible representation selection with persistent memory pricing to adapt allocation over time while decoupling semantic refresh from resource accounting.

\item We prove a linear worst-case regret lower bound against
independent-state hindsight and introduce a same-state benchmark.
Under stated assumptions, \textsc{MemPrice} admits regret bounds
with explicit estimation-error and priced-frontier terms, and
\(O(\sqrt{T})\) cumulative occupancy violation.

\item We evaluate memory allocation under two complementary settings: \emph{Query-Blind}, where future information needs are unknown during memory construction, and \emph{Goal-Observable}, where the task goal is known from the outset and can guide evidence accumulation. Our method achieves strong results in both settings.

\end{itemize}

\begin{figure}[t]
    \centering
    \includegraphics[width=\linewidth]{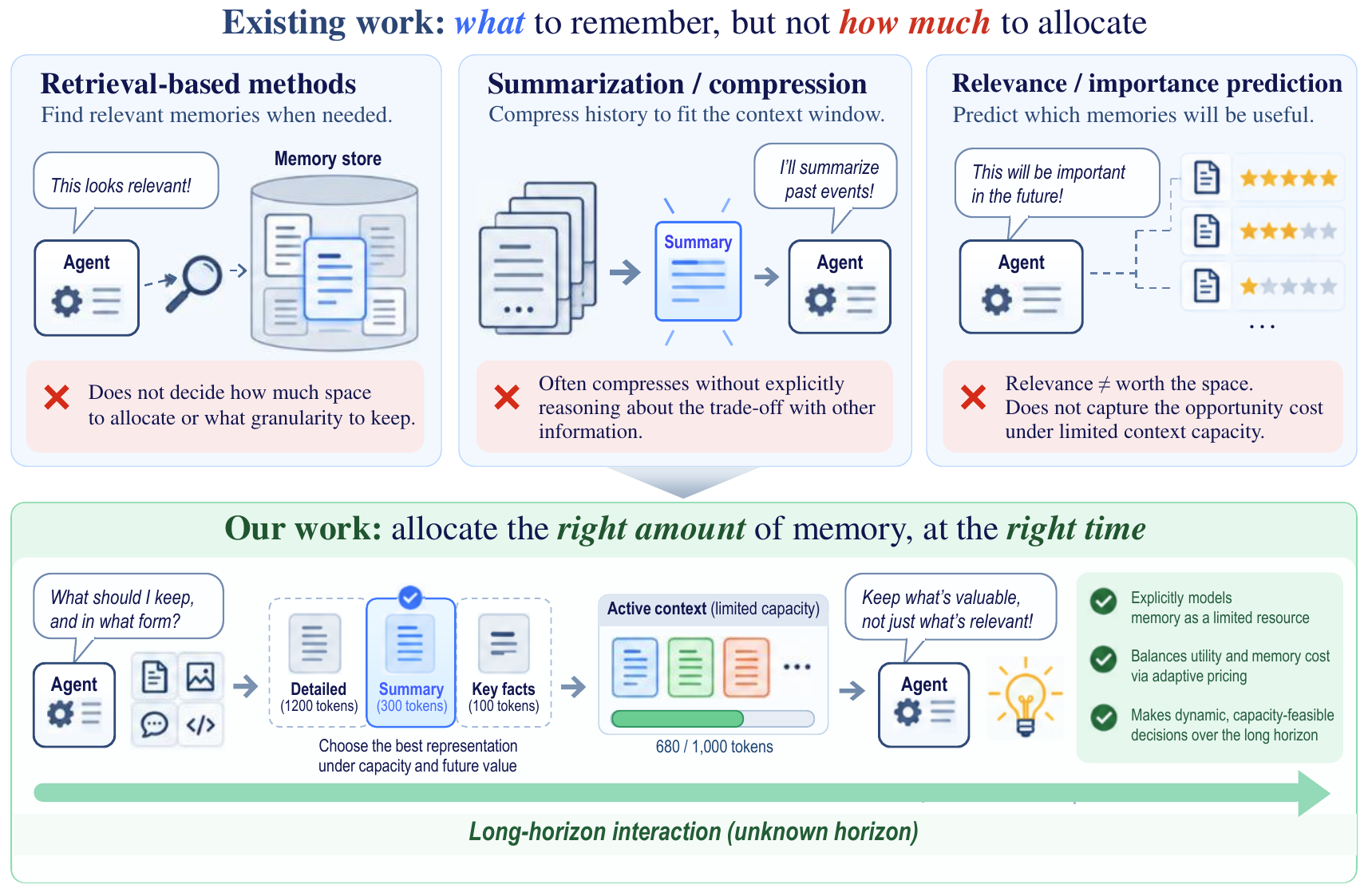}
    \setlength{\abovecaptionskip}{1pt}
    \caption{Comparison of existing works and ours.
    }
    \label{fig:memprice}
\end{figure}

\paragraph{Relation to prior work.} Long-horizon agents often rely on persistent memory to preserve and recover useful information over time \citep{shinn2023reflexion,park2023generative,packer2023memgpt, chhikara2025mem0,xu2025amem}. Recent work further adapts the contents and granularity of working context through trajectory folding and relevance-aware context construction \citep{sun2025contextfolding,ye2026agentfold,wei2026pace}. We view active context as a shared resource and formulate memory management as an online allocation problem across competing memories and representation levels, connecting agent memory to online resource allocation and long-term constrained optimization \citep{devanur2012online,agrawal2014dynamic,jenatton2016adaptive,yu2017online,zong2026online}. MemPrice realizes this view through adaptive pricing and persistent resource accounting across memory refreshes. A detailed discussion of related work is provided in Appendix~\ref{app:lr}.


\section{Problem Formulation}
\label{sec:problem-formulation}
\paragraph{Agent memory allocation.}
We study online allocation of an agent's active context over an interaction of unknown length. At each round, the allocator selects representations using only information available so far. Because compression may be irreversible, past decisions can restrict future feasible allocations. The horizon \(T\) denotes an arbitrary evaluation prefix and is unknown to the allocator.

Let \(B>0\) denote the hard active-context capacity, and let \(\alpha\in(0,1]\) specify the target average occupancy ratio. At round \(t\), the allocator chooses \(a_t\in\mathcal A_t(B)\), where \(\mathcal A_t(B)\) is the feasible set induced by the current memory state. Let \(U_t(a_t)\) and \(G_t(a_t)\) denote retention utility and memory occupancy, and define \(g_t(a_t)=G_t(a_t)/B\). For a non-anticipating policy \(\pi\), the finite-prefix reference objective is
\begin{equation}
\label{eq:online-memory-objective}
\begin{aligned}
\max_{\pi}\quad
\mathbb{E}_{\pi}\!\left[\sum_{t=1}^{T} U_t(a_t)\right] 
\mathrm{s.t.}\quad
a_t\in\mathcal{A}_t(B)\ \forall t,\quad
\mathbb{E}_{\pi}\!\left[\frac{1}{T}\sum_{t=1}^{T} g_t(a_t)\right]\le \alpha,
\end{aligned}
\end{equation}
The feasible sets \(\mathcal A_t(B)\) are induced by the memory-state trajectory generated by \(\pi\).

\paragraph{Memory representation choices.}
Let \(\mathcal{R}_t\) denote the set of memory items currently resident in active context at round \(t\). Each item can be retained at one of five representation levels,
\[
\mathrm{raw}
\succ
\mathrm{summary\,(sum)}
\succ
\mathrm{abstract\,(abs)}
\succ
\mathrm{keywords\,(key)}
\succ
\mathrm{deletion\,(del)}
\]
ordered from more detailed to more compressed representations. Let \(c_i^{-}(t)\) denote the representation of item \(i\) available immediately before the round-\(t\) allocation. Between memory refreshes, compression is one-way: the allocator may keep an item's current representation or move to a coarser level, but cannot recover discarded detail. For item \(i\) at representation level \(j\), let \(s_{i,t}^{(j)}\) denote its memory footprint. Coarser representations use no more memory, with deletion having zero footprint. For an allocation \(a=(c_i)_{i\in\mathcal{R}_t}\), its total memory occupancy is $G_t(a)=\sum_{i\in\mathcal{R}_t}s_{i,t}^{(c_i)}.$ The feasible allocation set in Eq.~\eqref{eq:online-memory-objective} is therefore
\[
    \mathcal{A}_t(B)
    =
    \left\{
        a=(c_i)_{i\in\mathcal{R}_t}
        :
        c_i\preceq c_i^{-}(t)\ \forall i\in\mathcal{R}_t,\;
        G_t(a)\le B
    \right\}.
\]

Let \(u_{i,t}^{(j)}\) denote the true retention utility of keeping item $i$ at representation \(j\). We assume that retention utility does not increase under further compression and vanishes upon deletion. At each round, an LLM oracle provides utility estimates \(\widehat{u}_{i,t}^{(j)}\) for the feasible representation choices. Let
\[
    U_t(a)
    =
    \sum_{i\in\mathcal{R}_t}u_{i,t}^{(c_i)},
    \qquad
    \widehat{U}_t(a)
    =
    \sum_{i\in\mathcal{R}_t}\widehat{u}_{i,t}^{(c_i)}.
\]
Throughout the analysis, \(U_t\) denotes the allocation-level retention objective; Section~5 evaluates how the resulting memory policies translate to downstream task performance.

\paragraph{Memory refresh.}
The interaction is organized into consecutive refresh windows of \(L\) rounds, except possibly for the final window. Within each window, the resident set \(\mathcal{R}_t\) consists of the context carried from the previous boundary together with newly observed interaction items. Motivated by persistent-memory designs that retain interaction history for later retrieval, we store raw transcripts in an external archive and use a fixed compression--retrieval pipeline at refresh boundaries to construct the carried context for the next window. Thus, compression is irreversible within a refresh window but archived information can be recovered at subsequent boundaries. Memory occupancy and retention utility are defined over the active context only.

\section{Algorithm}
\label{sec:revised-algorithm}
\begin{figure}[t]
    \centering
    \includegraphics[width=\linewidth]{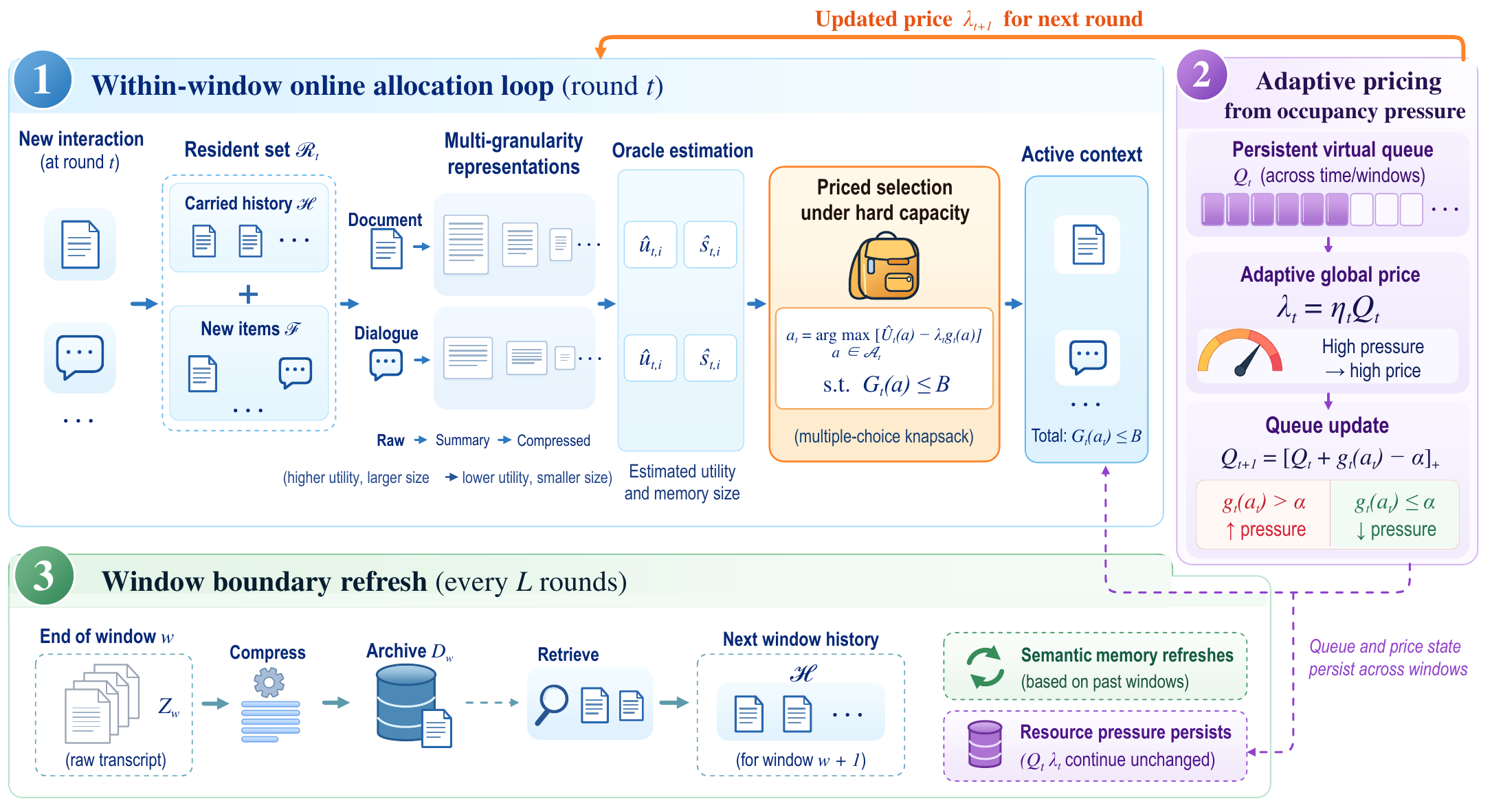}
    \setlength{\abovecaptionskip}{1pt}
    \caption{An overview of our MemPrice framework.
    }
    \label{fig:memprice}
\end{figure}

MemPrice controls active-context allocation through a history-dependent memory price. At each round, accumulated occupancy pressure determines the current memory price. The allocator then jointly selects a representation for each resident memory, retaining additional detail only when its estimated value justifies the corresponding context cost. This pressure carries across memory refreshes, allowing the agent to refresh its active context without forgetting how heavily context has been used. Algorithm~\ref{alg:memprice} summarizes the full procedure.

\paragraph{Adaptive memory pressure.}
Although the hard capacity $B$ prevents context overflow, enforcing a fixed fraction of $B$ at each round would be unnecessarily rigid: the active context at some rounds contains information that warrants substantially more detailed retention than at others. MemPrice therefore uses \(\alpha B\) as a long-run occupancy target, allowing temporary increases in context usage when additional memory detail is valuable. A virtual queue \(Q_t\) tracks accumulated occupancy pressure relative to this target and converts it into a memory price:
\[
\eta_t = \frac{c}{\sqrt{t}},~\lambda_t = \eta_t Q_t,~Q_{t+1}=\left[Q_t+g_t(a_t)-\alpha\right]_+,
\]
where $[x]_+=\max\{x,0\}$ and $c>0$ controls the strength of the pricing feedback: larger \(c\) makes the allocator respond more aggressively to accumulated occupancy pressure. Occupancy above the target increases \(Q_t\), while lighter usage gradually releases this pressure. The resulting memory price \(\lambda_t\) is shared across all resident memories: when pressure is low, detailed representations are relatively cheap; as pressure accumulates, retaining extra detail requires correspondingly higher estimated utility.

\paragraph{Price-aware joint allocation.}
Given the global price \(\lambda_t\), MemPrice selects the capacity-feasible representation assignment that maximizes a price-adjusted retention objective:
\begin{equation}
\label{eq:revised-priced-action}
    a_t
    \in
    \operatorname*{arg\,max}_{a\in\mathcal A_t(B)}
    \left\{
        \widehat U_t(a)-\lambda_t g_t(a)
    \right\}.
\end{equation}
The price controls how much detail is worth retaining rather than simply filtering memories by relevance. In particular, \(\lambda_t/B\) can be viewed as a per-token context price: additional detail is retained only when its estimated value justifies its extra footprint. Since all resident memories share the same context budget, their representation choices are coupled: a detailed representation may be worthwhile when competition for context is light, but not when more valuable information is competing for the same space. Therefore, Eq.~\eqref{eq:revised-priced-action} becomes a multiple-choice knapsack problem (MCKP), which jointly selects one admissible representation for each resident
memory:
\begin{equation}
\label{eq:revised-priced-mckp}
    (c_i(t))_{i\in \mathcal R_t}
    \in
    \operatorname*{arg\,max}_{(c_i)_{i\in \mathcal R_t}\in\mathcal A_t(B)}
    \sum_{i\in \mathcal R_t}
    \left(
        \widehat u_{i,t}^{(c_i)}
        -
        \frac{\lambda_t}{B}s_{i,t}^{(c_i)}
    \right).
\end{equation}

\paragraph{Persistent memory pressure across refreshes.}
At each refresh boundary, the raw transcript from the current window is added to the persistent archive, and a fixed compression--retrieval pipeline constructs the carried context for the next window. This refresh can reintroduce information that was previously compressed out of the active context, making earlier detail available again when needed. The refreshed context remains subject to the hard capacity \(B\), while the virtual queue \(Q_t\) and global round index $t$ are carried across the boundary.

\begin{algorithm}[t]
\caption{MemPrice: Online Memory Allocation with Adaptive Pricing}
\label{alg:memprice}
\small
\begin{algorithmic}[1]

\REQUIRE Hard capacity $B$, target occupancy ratio $\alpha$, pricing constant $c$, window length $L$

\STATE Initialize round $t\gets1$, pressure $Q_1\gets0$, carried context $H\gets\varnothing$, current-window items
$\mathcal F\gets\varnothing$

\WHILE{\textbf{true}}
    \FOR{$k=1,\ldots,L$}

        \STATE Observe a new interaction item $\theta_t$, add it to $\mathcal F$ at level $\mathrm{raw}$, and form the resident set $\mathcal R_t\gets\{H\}\cup\mathcal F$
        \STATE For each $i\in\mathcal R_t$, query the LLM oracle for admissible representations and utility estimates $\{\widehat u_{i,t}^{(j)}\}$; record their sizes $\{s_{i,t}^{(j)}\}$

        \STATE Set $\eta_t\gets c/\sqrt t$ and $\lambda_t\gets\eta_tQ_t$

        \STATE  Select $a_t$ by solving Eq. ~\ref{eq:revised-priced-mckp}

        \STATE Apply $a_t$ and update
        \[
            Q_{t+1}
            \gets
            \left[Q_t+g_t(a_t)-\alpha\right]_+
        \]

        \IF{the conversation ends}
        \STATE Archive the current-window interaction history and \textbf{terminate}
        \ENDIF

        \STATE $t\gets t+1$

    \ENDFOR

    \STATE Archive the current-window interaction history and perform \textsc{MemoryRefresh} to obtain the carried context $H$ for the next window; set $\mathcal F \gets \varnothing$

\ENDWHILE

\end{algorithmic}
\end{algorithm}

\section{Theoretical Analysis}
\label{sec:performance-guarantee}
Our analysis distinguishes two sources of hindsight advantage: access
to representations preserved along a different memory trajectory,
and the ability to redistribute occupancy across time under a global
budget. We first characterize the effect of compression-induced state
divergence, then evaluate allocation quality on MemPrice's realized
memory-state trajectory.
To formalize the first effect, consider an \emph{independent-state
hindsight comparator}: an optimal policy that observes the full
\(T\)-round interaction prefix and follows its own memory trajectory
under the same fixed refresh pipeline. Its allocations \(b_t\) are
subject to the hard capacity \(B\) at every round and the cumulative
occupancy budget \(\sum_{t=1}^T g_t(b_t)\le \alpha T\).
Let \(\operatorname{OPT}_T^{\mathrm{ind}}\) denote its optimal cumulative retention utility, and define
\[
\operatorname{Reg}_T^{\mathrm{ind}}
:=
\mathbb{E}\!\left[
\operatorname{OPT}_T^{\mathrm{ind}}
-
\sum_{t=1}^T U_t(a_t)
\right].
\]
The following proposition establishes a linear worst-case regret
lower bound for every capacity-feasible, non-anticipating allocator,
even with exact utility estimates.

\begin{proposition}[Linear regret under independent-state comparison]
\label{prop:independent-state-linear-regret}
Let \(B>0\), \(\alpha\in(0,1]\), and \(M:=\lfloor 1/\alpha\rfloor+1\). For any non-anticipating allocator with \(a_t\in\mathcal A_t(B)\) for all \(t\) and integers \(L,T>M\), there exists a \(T\)-round instance with exact utility estimates such that
\begin{equation}
\label{eq:independent-state-lower-bound}
    \operatorname{Reg}_T^{\mathrm{ind}}
    \ge
    \frac{T-M}{M(M+1)}.
\end{equation}
\end{proposition}


Proposition~\ref{prop:independent-state-linear-regret} identifies an information-loss barrier rather than an estimation barrier. Before future relevance is revealed, hard capacity forces the online allocator to discard some candidate detail, whereas hindsight preserves the candidate that will later matter. Importantly, the hindsight policy in the construction uses at most
\(\alpha B\) at every round, so the lower bound does not rely on
redistributing occupancy across time. To isolate the allocation component, we next introduce a same-state comparator that evaluates hindsight decisions on the realized feasible-state trajectory.


For each round \(t\), let \(\mathcal A_t^{\mathrm{alg}}(B)\) denote the feasible allocation set induced by MemPrice's realized memory state. We define the global same-state (GSS) hindsight benchmark over a \(T\)-round prefix as
\begin{equation}
\label{eq:same-state-hindsight-opt}
\operatorname{OPT}_{T}^{\mathrm{gss}}
:=
\max_{b_1,\ldots,b_T}
\sum_{t=1}^{T} U_t(b_t)
\quad \mathrm{s.t.}\quad
b_t\in\mathcal A_t^{\mathrm{alg}}(B),\ \forall t,\qquad
\sum_{t=1}^{T} g_t(b_t)\le\alpha T .
\end{equation}
The GSS benchmark performs counterfactual reallocation on MemPrice’s realized memory states. It retains full hindsight and may redistribute occupancy across the entire prefix, but its alternative choices do not propagate to future states. Thus, GSS regret measures allocation loss conditional on the states reached by MemPrice, rather than the long-term cost of generating those states.
The occupancy budget is imposed pathwise on the realized prefix, and Appendix~\ref{app:expected} gives the expectation-feasible counterpart aligned with Eq.~\ref{eq:online-memory-objective}.
\begin{equation}
\label{eq:allocation-regret-violation}
\operatorname{Reg}_{T}^{\mathrm{gss}}
=
\mathbb E\!\left[
    \operatorname{OPT}_{T}^{\mathrm{gss}}
    -
    \sum_{t=1}^{T}U_t(a_t)
\right],
\qquad
\operatorname{Viol}_{T}
=
B\left[
    \sum_{t=1}^{T}\bigl(g_t(a_t)-\alpha\bigr)
\right]_+ .
\end{equation}

To account for errors in the LLM oracle's retention-utility estimates, we make the following bounded-error assumption.

\begin{assumption}[Bounded utility and estimation error]
\label{ass:utility-oracle}
There exist finite constants $\overline U>0$ and $\overline\epsilon\ge0$
such that, for every $t$, almost surely,
\(0\le U_t(a)\le\overline U\) for all
\(a\in\mathcal A_t^{\mathrm{alg}}(B)\) and
\(e_t:=\sup_{a\in\mathcal A_t^{\mathrm{alg}}(B)}
|\widehat U_t(a)-U_t(a)|\le\overline\epsilon\).
\end{assumption}

We write \(\mathcal E_T:=\sum_{t=1}^T\mathbb E[e_t]\) for the expected accumulated estimation error. Under a global occupancy budget, hindsight may redistribute memory usage across time. To control this temporal advantage, define the priced frontier
$$
\Phi_t(\lambda)
:=
\max_{a\in\mathcal A_t^{\mathrm{alg}}(B)}
\left\{
U_t(a)-\lambda\bigl(g_t(a)-\alpha\bigr)
\right\}.
$$

\begin{assumption}[Priced-frontier regularity]
\label{ass:stability}
There exists a deterministic sequence $\Delta_T=o(T)$ such that, for every $T$, almost surely,
\[
\inf_{\lambda\ge0}
\sum_{t=1}^T
\left[
\Phi_t(\lambda)
-
\inf_{\mu\ge0}\Phi_t(\mu)
\right]
\le \Delta_T .
\]
\end{assumption}
Priced-frontier regularity requires the realized utility–occupancy frontiers to admit an approximately common minimizing price. In particular, \(\Delta_T=0\) whenever the round-wise frontiers share a minimizer, even if the resident memories and optimal allocations vary across rounds. Appendix~\ref{app:frontier-regularity} gives a sufficient condition allowing dispersion in these minimizing prices. This condition controls the remaining hindsight advantage from reallocating occupancy across time and is not required for hard-capacity feasibility or long-run occupancy control.
The following result gives the corresponding same-state allocation and occupancy guarantees.
\begin{theorem}[Same-state allocation and occupancy guarantees]
\label{thm:global-guarantee}
Algorithm~\ref{alg:memprice} maintains
\(G_t(a_t)\le B\) at every round by construction.
Under Assumptions~\ref{ass:utility-oracle} and~\ref{ass:stability},
for any evaluation prefix \(T\ge1\),
\begin{equation}
\label{eq:global-performance-bounds}
\operatorname{Reg}_{T}^{\mathrm{gss}}
\le
c\sqrt T+2\mathcal E_T+\Delta_T,
\qquad
\operatorname{Viol}_{T}
\le
B\left(
\frac{\overline U+2\overline\epsilon}{\alpha c}\sqrt T+1
\right)
\quad\text{a.s.}
\end{equation}
\end{theorem}



Theorem~\ref{thm:global-guarantee} establishes same-state allocation and occupancy guarantees on MemPrice's realized memory-state trajectory. MemPrice never exceeds the hard capacity \(B\), whereas \(\alpha B\) acts as a long-run operating target rather than a per-round cap; consequently, the excess average occupancy vanishes at rate \(O(T^{-1/2})\).
The regret bound decomposes into the online pricing term \(c\sqrt T\), the accumulated estimation error \(\mathcal E_T\), and the priced-frontier deviation \(\Delta_T\). For fixed \(c\), \(\mathcal E_T+\Delta_T=o(T)\) yields vanishing average GSS regret, while \(\mathcal E_T,\Delta_T=O(\sqrt T)\) gives \(O(\sqrt T)\) cumulative GSS regret. Larger \(c\) strengthens occupancy feedback, tightening the \(O(\sqrt T/c)\) occupancy bound at the cost of a larger \(O(c\sqrt T)\) online pricing term.

\section{Experiments}
\label{sec:experiment}
\subsection{Experimental Settings}
\label{subsec:exp-setup}
\newcommand{\res}[1]{%
  $#1$%
}
\newcommand{\bestres}[1]{%
  $\mathbf{#1}$%
}
\newcommand{\secondres}[1]{%
  $\underline{#1}$%
}

\begin{table*}[!t]
\centering
\caption{
\textbf{Results for Setting I} under matched hard memory budgets.
All budget-constrained methods use the same persistent-memory budget, retrieval protocol, and answer model within each backbone block. Best and second-best results among budget-constrained methods are shown in \textbf{bold} and \underline{underline}, respectively.
Full Context$^\dagger$ is an unconstrained reference and is excluded
from ranking.
}
\label{tab:main-experiment-i}

\scriptsize
\renewcommand{\arraystretch}{0.96}
\setlength{\tabcolsep}{1.7pt}

\begin{tabular*}{\textwidth}{
@{\extracolsep{\fill}}
l
ccccc
cccc
@{}
}
\toprule
&
\multicolumn{5}{c}{LongMemEval-S ($B=16\mathrm{k}$), QA Acc. (\%)}
&
\multicolumn{4}{c}{LoCoMo ($B=8\mathrm{k}$), F1 (\%)} \\
\cmidrule(lr){2-6}
\cmidrule(l){7-10}

Method
& \shortstack{S-Hop\\{\tiny $n=156$}}
& \shortstack{M-Hop\\{\tiny $n=133$}}
& \shortstack{Temp.\\{\tiny $n=133$}}
& \shortstack{K-Upd.\\{\tiny $n=78$}}
& \shortstack{Overall\\{\tiny $n=500$}}
& \shortstack{S-Hop\\{\tiny $n=841$}}
& \shortstack{M-Hop\\{\tiny $n=282$}}
& \shortstack{Temp.\\{\tiny $n=321$}}
& \shortstack{Overall\\{\tiny $n=1444$}} \\
\midrule


\multicolumn{10}{@{}l}{\textit{GLM-5.3-Flash}} \\

Full Context$^\dagger$
& \res{94.9}
& \res{80.5}
& \res{89.5}
& \res{87.2}
& \res{88.4}
& \res{62.5}
& \res{48.8}
& \res{53.3}
& \res{57.8} \\

Recent / FIFO
& \res{43.2}
& \res{24.7}
& \res{29.1}
& \res{39.8}
& \res{34.0}
& \res{35.2}
& \res{23.7}
& \res{27.1}
& \res{31.2} \\

Uniform Allocation
& \res{54.1}
& \res{39.8}
& \res{45.2}
& \res{40.6}
& \res{45.8}
& \res{39.4}
& \res{30.2}
& \res{29.7}
& \res{35.4} \\

Content Salience
& \secondres{78.6}
& \secondres{64.3}
& \secondres{72.8}
& \secondres{68.9}
& \secondres{71.7}
& \res{42.8}
& \res{35.9}
& \res{34.7}
& \res{39.7} \\

Dense Salience
& \res{72.3}
& \res{59.4}
& \res{65.1}
& \res{62.7}
& \res{65.5}
& \res{46.6}
& \res{37.8}
& \res{40.5}
& \res{43.5} \\

LRE-SelfSup
& \res{75.2}
& \res{61.7}
& \res{69.8}
& \res{66.4}
& \res{68.8}
& \bestres{49.9}
& \secondres{41.3}
& \secondres{43.0}
& \secondres{46.7} \\

\textbf{MemPrice (Ours)}
& \bestres{80.0}
& \bestres{66.8}
& \bestres{79.6}
& \bestres{77.3}
& \bestres{76.0}
& \secondres{49.4}
& \bestres{41.9}
& \bestres{48.8}
& \bestres{47.8} \\

\addlinespace[2pt]

\multicolumn{10}{@{}l}{\textit{DeepSeek-V4.1-Flash}} \\

Full Context$^\dagger$
& \res{92.8}
& \res{81.8}
& \res{87.4}
& \res{88.1}
& \res{87.7}
& \res{58.9}
& \res{52.0}
& \res{48.7}
& \res{55.3} \\

Recent / FIFO
& \res{40.5}
& \res{27.1}
& \res{26.8}
& \res{41.2}
& \res{33.4}
& \res{31.9}
& \res{25.2}
& \res{26.6}
& \res{29.4} \\

Uniform Allocation
& \res{52.6}
& \res{44.1}
& \res{42.8}
& \res{42.2}
& \res{46.1}
& \res{37.5}
& \res{28.1}
& \res{31.8}
& \res{34.4} \\

Content Salience
& \secondres{76.1}
& \secondres{66.2}
& \secondres{70.4}
& \secondres{68.7}
& \secondres{70.8}
& \res{40.9}
& \res{34.7}
& \res{39.0}
& \res{39.3} \\

Dense Salience
& \res{69.2}
& \res{60.5}
& \res{64.3}
& \res{59.8}
& \res{64.1}
& \res{44.2}
& \res{39.1}
& \bestres{45.3}
& \res{43.4} \\

LRE-SelfSup
& \res{73.5}
& \res{63.4}
& \res{68.7}
& \res{66.9}
& \res{68.5}
& \secondres{47.2}
& \secondres{40.6}
& \secondres{44.8}
& \secondres{45.4} \\

\textbf{MemPrice (Ours)}
& \bestres{77.4}
& \bestres{73.0}
& \bestres{73.4}
& \bestres{75.1}
& \bestres{74.8}
& \bestres{47.3}
& \bestres{45.0}
& \res{44.1}
& \bestres{46.1} \\

\addlinespace[2pt]

\multicolumn{10}{@{}l}{\textit{Qwen3.8-Flash}} \\

Full Context$^\dagger$
& \res{95.4}
& \res{82.0}
& \res{90.8}
& \res{88.2}
& \res{89.5}
& \res{63.7}
& \res{49.9}
& \res{55.8}
& \res{59.2} \\

Recent / FIFO
& \res{42.4}
& \res{28.3}
& \res{26.1}
& \res{34.5}
& \res{33.1}
& \res{34.8}
& \res{24.7}
& \res{24.5}
& \res{30.5} \\

Uniform Allocation
& \res{55.3}
& \res{38.9}
& \res{49.0}
& \res{50.2}
& \res{48.5}
& \res{38.1}
& \res{29.6}
& \res{33.0}
& \res{35.3} \\

Content Salience
& \secondres{78.2}
& \secondres{63.9}
& \secondres{73.5}
& \secondres{68.4}
& \secondres{71.6}
& \res{43.5}
& \res{34.2}
& \res{36.7}
& \res{40.2} \\

Dense Salience
& \res{72.5}
& \res{59.3}
& \res{68.2}
& \res{66.8}
& \res{67.0}
& \bestres{49.1}
& \res{38.5}
& \res{41.9}
& \res{45.4} \\

LRE-SelfSup
& \res{75.4}
& \res{62.1}
& \res{71.0}
& \res{67.9}
& \res{69.5}
& \secondres{48.8}
& \secondres{42.7}
& \secondres{42.0}
& \secondres{46.1} \\

\textbf{MemPrice (Ours)}
& \bestres{80.4}
& \bestres{67.5}
& \bestres{80.2}
& \bestres{72.0}
& \bestres{75.6}
& \res{48.5}
& \bestres{43.0}
& \bestres{48.1}
& \bestres{47.3} \\

\bottomrule
\end{tabular*}

\vspace{1pt}
\end{table*}
\paragraph{Experimental settings.}
We evaluate memory allocation under two information regimes.
\textbf{Setting I} is \emph{query-blind}: the terminal query is hidden while
memory is constructed, and the allocator can use only the chronological
history observed so far. We evaluate on
\textbf{LongMemEval-S}~\citep{wu2025longmemeval} and
\textbf{LoCoMo}~\citep{maharana2024locomo}.
We compare with query-blind controls including
\textbf{Recent/FIFO}, \textbf{Uniform Allocation},
\textbf{Content Salience}, \textbf{Dense Salience}, and
\textbf{LRE-SelfSup}~\citep{lia2026lre}.
\textbf{Full Context} is reported separately as an unbudgeted reference.
\textbf{Setting II} is \emph{goal-observable}: the task goal is visible from
reset and may guide allocation, while future observations, tool outputs, and
evaluator information remain hidden. We evaluate on
\textbf{AppWorld}~\citep{trivedi2024appworld} and
\textbf{WebShop}~\citep{yao2022webshop}, comparing against
\textbf{ReAct/FIFO History}~\citep{yao2023react},
a \textbf{Summary Agent}, and \textbf{ZipAct-B}, our budget-matched adaptation
of ZipAct~\citep{pan2026zipact}.
All budgeted methods share the same tokenizer-measured historical-memory
capacity, information contract, task split, evaluator, and failure denominator.

We repeat the main comparison with \textbf{GLM-5.3-Flash}~\cite{glm5team2026glm5vibecodingagentic}, \textbf{Qwen3.8-Flash}~\cite{qwen2026design}, and \textbf{DeepSeek-V4.1-Flash} ~\cite{deepseekai2026deepseekv41flash}.
For each model--dataset pair, all methods use the same reader/agent,
prompting interface, tokenizer, and decoding configuration.
The historical-memory budget \(B\) excludes the current query or goal,
current observation, system prompt, tool schemas, and output reserve.
Our method uses five ordered representations---raw, summary, abstraction,
keywords, and delete---and jointly allocates them with an exact
multiple-choice knapsack solver.
We set the target occupancy to \(\alpha=0.8\), maintain the persistent pressure
state
\(\lambda_t=cQ_t/\sqrt{t}\), and trigger maintenance at capacity pressure or
after at most \(L=10\) interactions.
The scale \(c\) is selected on development data using occupancy only.
We report the primary comparison at \(B^\star=B_{\max}/2\) and average all main results over three random seeds.

\subsection{Main Results}
\label{subsec:exp-main}
\paragraph{Setting I: query-blind resource allocation}
Table~\ref{tab:main-experiment-i} shows a consistent advantage of MemPrice under matched memory budgets. 
$^\dagger$Full Context uses the complete history and is reported only as an unconstrained reference; it is excluded from best/second-best ranking.
All reported Overall scores are weighted by the corresponding category sizes.
On LongMemEval-S, MemPrice improves the strongest constrained baseline across the three backbones and ranks first in every query category. More importantly, the gain is highly non-uniform across query types: improvements are modest on single-hop questions, but become substantially larger on multi-hop, temporal, and knowledge-update queries. This pattern suggests that MemPrice is most useful when memory items compete for capacity over longer dependencies, where simple recency or salience scores are insufficient to capture their future value. LoCoMo exhibits the same trend in a weaker form: MemPrice remains best overall across all backbones, while individual single-hop or temporal subsets can still favor LRE-SelfSup or Dense Salience. This indicates that the benefit comes primarily from difficult allocation decisions rather than uniformly improving retrieval. Finally, MemPrice still trails Full Context, showing that better allocation cannot fully recover information removed by a hard budget; instead, it narrows the performance loss by preserving a more useful subset of the history.

\begin{table}[!t]
\centering
\caption{
\textbf{Results for Setting II} in the goal-observable setting at
$B_{\mathrm{main}}=8$K historical-memory tokens.
All budget-constrained methods use the same persistent-memory budget,
retrieval protocol, and answer model within each backbone block.
Reported results are mean $\pm$ SD over three independent seeds.
}
\label{tab:main-experiment-ii}
\small
\setlength{\tabcolsep}{5pt}

\begin{tabular}{llccc}
\toprule
Model & Method
& \multicolumn{2}{c}{AppWorld TGC (\%)}
& WebShop SR (\%) \\
\cmidrule(lr){3-4}
& & Normal & Challenge & \\
\midrule

\multirow{4}{*}{GLM-5.3-Flash}
& ReAct / FIFO History
& $35.1\pm1.9$
& $19.3\pm1.7$
& $43.6\pm1.5$ \\
& Summary Agent
& $33.8\pm2.1$
& $18.1\pm1.9$
& $40.8\pm1.7$ \\
& ZipAct
& $\underline{39.8\pm1.5}$
& $\underline{23.7\pm1.4}$
& $\underline{49.6\pm1.3}$ \\
& \textbf{MemPrice (Ours)}
& $\mathbf{42.2\pm1.7}$
& $\mathbf{25.8\pm1.5}$
& $\mathbf{53.1\pm1.4}$ \\
\midrule

\multirow{4}{*}{DeepSeek-V4.1-Flash}
& ReAct / FIFO History
& $36.9\pm1.9$
& $20.9\pm1.7$
& $45.8\pm1.6$ \\
& Summary Agent
& $35.3\pm1.6$
& $19.4\pm2.0$
& $43.1\pm1.8$ \\
& ZipAct
& $\underline{41.4\pm1.4}$
& $\underline{25.0\pm1.5}$
& $\underline{52.0\pm1.5}$ \\
& \textbf{MemPrice (Ours)}
& $\mathbf{44.0\pm1.6}$
& $\mathbf{27.0\pm1.8}$
& $\mathbf{55.8\pm1.3}$ \\
\midrule

\multirow{4}{*}{Qwen3.8-Flash}
& ReAct / FIFO History
& $36.0\pm1.7$
& $20.1\pm1.5$
& $46.2\pm1.5$ \\
& Summary Agent
& $34.7\pm1.9$
& $18.7\pm1.7$
& $43.5\pm1.9$ \\
& ZipAct
& $\underline{40.4\pm1.6}$
& $\underline{24.1\pm1.3}$
& $\underline{50.8\pm1.4}$ \\
& \textbf{MemPrice (Ours)}
& $\mathbf{42.7\pm1.5}$
& $\mathbf{26.3\pm1.6}$
& $\mathbf{55.1\pm1.6}$ \\
\bottomrule
\end{tabular}

\end{table}
\paragraph{Setting II: goal-observable resource allocation}
Table~\ref{tab:main-experiment-ii} shows that MemPrice consistently outperforms all baselines across three backbones and both agent benchmarks, with ZipAct being the strongest competitor in every setting. The improvement is particularly pronounced on WebShop, where MemPrice exceeds ZipAct by \(2.7\)--\(3.3\) points, compared with more moderate gains on AppWorld. This suggests that adaptive memory allocation is especially beneficial for long-horizon interactive tasks, where preserving goal-relevant states and past actions is critical for subsequent decisions. In contrast, the smaller gains on AppWorld Challenge indicate that, as task difficulty increases, failures are increasingly driven by planning and execution rather than memory alone. We also observe that Summary Agent consistently underperforms even FIFO history, suggesting that aggressive summarization can discard fine-grained actionable information that remains important for agent execution. Overall, the consistent gains across backbones indicate that MemPrice improves how a fixed memory budget is allocated rather than relying on a particular underlying model.

\subsection{Ablation Study}
\label{subsec:exp-ablation}
\begin{table*}[t]
    \centering
    \small
    \setlength{\tabcolsep}{7pt}
    \renewcommand{\arraystretch}{1.08}
    \caption{
    Ablation study with GLM-5.3-Flash as the backbone.
    Each variant removes or replaces a single component of the full method
    while keeping all other components unchanged.
    FIFO History is included as a non-ablation reference.
    }
    \label{tab:ablation}

    \begin{tabular}{lcccc}
        \toprule
        Method
        & \shortstack{LongMemEval-S\\QA Acc. (\%)}
        & \shortstack{LoCoMo\\F1 (\%)}
        & \shortstack{AppWorld\\TGC (\%)}
        & \shortstack{WebShop\\SR (\%)} \\
        \midrule

        \textbf{Ours (Full)}
            & \textbf{76.0}
            & \textbf{47.8}
            & \textbf{42.2}
            & \textbf{53.1} \\

        Fixed Price
            & 70.9
            & 44.6
            & 41.5
            & 50.9 \\

        Raw/Delete
            & 34.7
            & 32.6
            & 36.3
            & 45.1 \\

        Greedy Allocation
            & 73.5
            & 45.8
            & 40.8
            & 51.7 \\

        Reset $Q$
            & 71.6
            & 45.2
            & 41.7
            & 50.5 \\

        No Recovery
            & 74.4
            & 44.1
            & 41.8
            & 50.7 \\

        \midrule

        FIFO History
            & 30.4
            & 31.2
            & 35.1
            & 43.6 \\

        \bottomrule
    \end{tabular}
\end{table*}
Table~\ref{tab:ablation} reveals several complementary roles of our design. The largest degradation comes from \textit{Raw/Delete}, especially on LongMemEval-S and LoCoMo, showing that memory management should not be reduced to deciding what to keep or discard; intermediate-fidelity representations are essential for preserving useful evidence under tight budgets. \textit{Fixed Price} and \textit{Reset $Q$} also hurt more on memory-intensive benchmarks, suggesting that memory pressure is both task-dependent and persistent over time, and therefore benefits from adaptive pricing and a stateful pressure signal. Replacing joint allocation with a greedy rule causes a smaller but consistent loss, indicating that coordination across memory items matters beyond local decisions. Finally, \textit{No Recovery} remains competitive on AppWorld but degrades more on LoCoMo and WebShop, suggesting that recovery is particularly useful when earlier compression decisions can become important again later in a long interaction. Overall, the ablations support our view that effective memory management requires not only compression, but adaptive and reversible allocation under evolving memory pressure.

\subsection{Sensitivity Analysis}
\label{subsec:sensitivity}
\begin{table*}[t]
\centering
\caption{
Sensitivity to the target occupancy parameter $\alpha$ at $B^\star$.
Task columns report each benchmark's primary quality metric;
for AppWorld, we report TGC on Test-Normal.
Mean occupancy is the normalized historical-memory load averaged across
the four benchmarks.
The default is $\alpha=0.8$.
}
\label{tab:sensitivity}
\small
\setlength{\tabcolsep}{7pt}
\renewcommand{\arraystretch}{1.05}

\begin{tabular*}{\textwidth}{@{\extracolsep{\fill}}cccccc}
\toprule
$\alpha$
& \shortstack{LongMemEval-S\\QA Acc. (\%)}
& \shortstack{LoCoMo\\F1 (\%)}
& \shortstack{AppWorld\\TGC (\%)}
& \shortstack{WebShop\\SR (\%)}
& \shortstack{Mean\\Occupancy} \\
\midrule

0.6
& 73.2
& 45.6
& 40.7
& 51.4
& 0.59 \\

0.7
& 75.1
& 47.2
& 41.8
& 52.7
& 0.70 \\

0.8
& \textbf{76.0}
& 47.8
& 42.2
& \textbf{53.1}
& 0.79 \\

0.9
& 75.8
& \textbf{48.0}
& \textbf{42.3}
& 53.0
& 0.88 \\

\bottomrule
\end{tabular*}
\end{table*}
Table~\ref{tab:sensitivity} studies the target occupancy parameter $\alpha$, which determines how aggressively the memory allocator frees capacity when memory pressure is encountered.
Increasing $\alpha$ from $0.6$ to $0.8$ consistently improves task quality across all four benchmarks.
With a small $\alpha$, the allocator leaves substantial unused headroom after compression, which reduces future memory pressure but also removes or downsamples useful historical information too aggressively.
Allowing the memory to operate at a higher occupancy preserves more relevant evidence and therefore improves downstream performance.
Importantly, $\alpha$ also controls a systems-level trade-off between memory utilization and inference cost.
A larger $\alpha$ allows the model to use a greater fraction of the available context but leaves less free capacity after each compression event.
As new observations accumulate, the memory therefore reaches the allocation boundary again sooner or compress tighter, leading to more frequent compression and reallocation operations.
These additional memory-management calls increase auxiliary token consumption and can also increase end-to-end inference latency.
Conversely, a smaller $\alpha$ reduces this overhead by maintaining more headroom but sacrifices task quality through more aggressive compression.

\section{Conclusions}
\label{sec:conclusion}
We study long-term agent memory as a resource-allocation problem under limited memory budgets. Rather than treating memory management as only retention, deletion, or compression, we consider how an agent should allocate limited capacity across historical information with different utility and storage costs. We propose a budget-aware framework that combines adaptive memory pricing, multi-fidelity representations, and joint allocation. Experiments on long-context QA and interactive agent benchmarks show that explicitly managing memory pressure leads to more effective and robust use of limited memory.
Our work has several limitations. The current framework relies on estimated utility signals and a predefined set of memory operations, and mainly focuses on memory capacity rather than jointly considering computation, retrieval, and latency. 


\bibliography{iclr2027/iclr2027_conference}
\bibliographystyle{iclr2027/iclr2027_conference}

\appendix
\section{Related Works}
\label{app:lr}
\paragraph{Agent memory and long-horizon context management.}
A growing body of work equips language agents with persistent memory beyond the model's immediate context window. Reflexion stores linguistic feedback from previous trials in an episodic memory buffer to improve subsequent decisions \citep{shinn2023reflexion}, while Generative Agents maintain a stream of experiences and retrieve memories according to factors such as relevance, recency, and importance, together with higher-level reflection \citep{park2023generative}. MemoryBank further introduces selective reinforcement and forgetting for long-term conversational memory \citep{zhong2023memorybank}. Other systems emphasize memory architecture and organization. MemGPT manages information across hierarchical memory tiers through an operating-system-inspired virtual context mechanism \citep{packer2023memgpt}; Mem0 dynamically extracts, consolidates, and retrieves salient conversational information \citep{chhikara2025mem0}; and A-MEM organizes memories into an evolving network whose representations and associations can change as new experiences arrive \citep{xu2025amem}. These approaches substantially extend the temporal reach of LLM agents, but primarily focus on how memories are stored, organized, updated, or retrieved. In contrast, our work isolates a complementary question: given the information currently available to an agent, how should scarce active-context capacity be allocated among competing memories and representation levels?

\paragraph{Adaptive context selection and compression.}
Recent work has moved beyond static memory retrieval toward actively controlling the contents and granularity of an agent's working context. Context-Folding allows an agent to branch into temporary sub-trajectories and subsequently fold intermediate reasoning into compact summaries, thereby preventing context growth on long-horizon tasks \citep{sun2025contextfolding}. AgentFold similarly treats the context as a dynamic workspace and learns proactive folding operations at different granularities, explicitly addressing the tension between unbounded raw trajectories and irreversible information loss from fixed summarization \citep{ye2026agentfold}. Most closely related to our setting, PACE formulates context construction as a Next Step Prediction problem and adjusts the representation granularity of historical information according to its predicted relevance for the agent's upcoming action \citep{wei2026pace}. These methods demonstrate the importance of selectively preserving detail rather than applying uniform truncation or summarization. Our perspective is complementary: rather than primarily predicting which history will be useful next or learning when to fold a trajectory, we explicitly model active context as a shared resource. Memory items with heterogeneous retention value and footprint compete for this resource, and allocation jointly determines both which items remain active and the granularity at which they are represented. This makes the opportunity cost of memory occupancy explicit and enables conversation-level resource control rather than treating context pressure only as a local selection or compression signal.

\paragraph{Online Resource Allocation and Long-Term Constraints.}
Online resource allocation can be viewed as sequential optimization under intertemporal constraints, where current decisions affect both immediate rewards and future resource availability. A classical approach is to associate resource constraints with dual prices that quantify the marginal value of capacity \citep{devanur2012online}. These prices can be updated dynamically as information accumulates \citep{agrawal2014dynamic,li2022online}, and related models have been studied under stochastic resource consumption \citep{jiang2020online} and non-stationary input distributions \citep{jiang2025online}. Such constrained allocation problems are closely connected to online optimization with long-term constraints, where primal-dual mechanisms balance objective performance and cumulative feasibility. This perspective has been studied in OCO through regret and constraint-violation guarantees \citep{jenatton2016adaptive,yu2017online,ao2026learning,zong2026online}, and also arises in resource-constrained learning problems such as bandits with knapsacks \citep{liu2022non}. 

\section{Expectation-feasible global same-state benchmark}
\label{app:expected}

The main text evaluates MemPrice against a global same-state comparator whose occupancy budget is enforced on each realized evaluation prefix. Here we consider a more permissive expectation-feasible counterpart that
matches the average-occupancy constraint in Eq.~\ref{eq:online-memory-objective}.
Let $\omega$ denote a realization of all randomness in the interaction and in MemPrice, including any randomness in the oracle and refresh pipeline. For each round $t$ and realization $\omega$, let
\[
    \mathcal A_t^{\mathrm{alg}}(B;\omega)
\]
denote the feasible allocation set induced by MemPrice's realized memory state. The memory-state trajectory is fixed to that generated by MemPrice: choices made by the hindsight comparator are evaluated counterfactually and do not alter the feasible sets at later rounds.

Let \(\zeta\) denote an auxiliary random seed independent of the randomness generating MemPrice's trajectory. A randomized same-state hindsight selector \(\pi\) observes the complete realized prefix and selects

$$ b_t^\pi(\omega,\zeta) \in \mathcal A_t^{\mathrm{alg}}(B;\omega). $$

Expectations below are taken jointly over \(\omega\) and \(\zeta\).
Thus, $\pi$ may use full hindsight when choosing each $b_t^\pi$, but it cannot recover representations that were unavailable on MemPrice's realized trajectory or induce a different future memory state. Let
$\Pi_T^{\mathrm{ss}}$ denote the set of all such measurable selectors.
We call a selector $\pi\in\Pi_T^{\mathrm{ss}}$ \emph{expectation-feasible} if
\begin{equation}
\label{eq:expected-feasibility}
    \mathbb E_{\omega,\zeta}\!\left[
        \sum_{t=1}^T g_t(b_t^\pi)
    \right]
    \le
    \alpha T.
\end{equation}
The per-round feasible-set constraint holds pathwise, whereas the cumulative occupancy budget in Eq.~\eqref{eq:expected-feasibility} is imposed only in expectation. Hence, the comparator may use more than $\alpha T$ total normalized occupancy on some realizations and less on others, while remaining within the hard capacity at every round.

Define
\[
    \Pi_T^{\mathrm{egss}}
    :=
    \left\{
        \pi\in\Pi_T^{\mathrm{ss}}:
        \mathbb E_{\omega,\zeta}\!\left[
            \sum_{t=1}^T g_t(b_t^\pi)
        \right]
        \le \alpha T
    \right\}.
\]
The expectation-feasible global same-state hindsight value and its associated
regret are
\begin{equation}
\label{eq:expected-gss-opt}
    \operatorname{OPT}_{T}^{\mathrm{egss}}
    :=
    \sup_{\pi\in\Pi_T^{\mathrm{egss}}}
    \mathbb E_{\omega,\zeta}\!\left[
        \sum_{t=1}^T U_t(b_t^\pi)
    \right],
\end{equation}
and
\begin{equation}
\label{eq:expected-gss-regret}
    \operatorname{Reg}_{T}^{\mathrm{egss}}
    :=
    \operatorname{OPT}_{T}^{\mathrm{egss}}
    -
    \mathbb E_{\omega,\zeta}\!\left[
        \sum_{t=1}^T U_t(a_t)
    \right].
\end{equation}
We refer to this comparator as EGSS. Every pathwise-feasible same-state selector is expectation-feasible, so this comparator is at least as permissive as the benchmark in the main text.

Since the cumulative occupancy constraint is imposed only in expectation, the corresponding dual comparison requires a deterministic shadow price shared across realizations.
\begin{assumption}[Expected priced-frontier regularity]
\label{ass:expected-stability}
For every \(T\), there exists a deterministic
\(\lambda_T^\star\ge0\) such that
\[
\mathbb E\!\left[
\sum_{t=1}^T
\left(
\Phi_t(\lambda_T^\star)
-
\inf_{\lambda\ge0}\Phi_t(\lambda)
\right)
\right]
\le
\Delta_T^{\mathrm E},
\]
where \(\Delta_T^{\mathrm E}=o(T)\) is deterministic.
\end{assumption}

\begin{theorem}[Expectation-level global guarantee]
\label{thm:expected-global-guarantee}
Under Assumptions~\ref{ass:utility-oracle}
and~\ref{ass:expected-stability}, for any evaluation prefix \(T\ge1\),
\begin{equation}
\label{eq:expected-global-regret}
\operatorname{Reg}_{T}^{\mathrm{egss}}
\le
c\sqrt T+2\mathcal E_T+\Delta_T^{\mathrm E}.
\end{equation}
Moreover, MemPrice satisfies
\begin{equation}
\label{eq:expected-average-occupancy}
\mathbb E\!\left[
\frac1T\sum_{t=1}^T g_t(a_t)
\right]
\le
\alpha
+
\frac{\overline U+2\overline\epsilon}{\alpha c\sqrt T}
+
\frac1T .
\end{equation}
\end{theorem}

Theorem~\ref{thm:expected-global-guarantee} shows that the main guarantees
extend to the more permissive expectation-feasible comparator aligned with
Eq.~\eqref{eq:online-memory-objective}. If
\(\mathcal E_T+\Delta_T^{\mathrm E}=o(T)\), the average EGSS regret vanishes;
if both terms are \(O(\sqrt T)\), the cumulative regret remains
\(O(\sqrt T)\). Moreover, the excess expected average occupancy above
\(\alpha\) vanishes as \(O(T^{-1/2})\).

Importantly, this extension relaxes only how the long-run occupancy budget is
enforced. The comparator remains same-state: it may redistribute context
usage across time and stochastic realizations, but cannot undo MemPrice's
earlier compression decisions or follow an independent memory trajectory.

\section{A Sufficient Condition for Priced-Frontier Regularity}
\label{app:frontier-regularity}

We give a simple sufficient condition for
Assumption~\ref{ass:stability} in terms of the dispersion of the
round-wise optimal shadow prices.

For each round \(t\), define
\[
\Lambda_t^\star
:=
\arg\min_{\lambda\ge0}\Phi_t(\lambda).
\]
The set \(\Lambda_t^\star\) is nonempty and contained in
\([0,\overline U/\alpha]\). Indeed, the all-delete action gives
\(\Phi_t(\lambda)\ge \alpha\lambda\), while
\(\Phi_t(0)\le\overline U\).
Moreover, since \(g_t(a)\in[0,1]\), the function
\(\Phi_t\) is \(1\)-Lipschitz in \(\lambda\). Therefore, for any
\(\lambda\ge0\),
\[
\Phi_t(\lambda)
-
\inf_{\mu\ge0}\Phi_t(\mu)
\le
\operatorname{dist}(\lambda,\Lambda_t^\star),
\]
where
\[
\operatorname{dist}(\lambda,S)
:=
\inf_{\mu\in S}|\lambda-\mu|.
\]

Consequently, Assumption~\ref{ass:stability} holds whenever there exists
a deterministic sequence \(R_T=o(T)\) such that, almost surely,
\[
\inf_{\lambda\in[0,\overline U/\alpha]}
\sum_{t=1}^T
\operatorname{dist}(\lambda,\Lambda_t^\star)
\le R_T.
\]

A more concrete sufficient condition is obtained when, for each prefix
\(T\), there exists a common price
\(\bar\lambda_T\in[0,\overline U/\alpha]\) such that all but
\(m_T=o(T)\) rounds satisfy
\[
\operatorname{dist}
(\bar\lambda_T,\Lambda_t^\star)
\le \rho_T,
\qquad
\rho_T=o(1).
\]
Then
\[
\inf_{\lambda\in[0,\overline U/\alpha]}
\sum_{t=1}^T
\operatorname{dist}(\lambda,\Lambda_t^\star)
\le
(T-m_T)\rho_T
+
m_T\frac{\overline U}{\alpha}
=
o(T),
\]
and hence Assumption~\ref{ass:stability} follows.
Thus, priced-frontier regularity permits a sublinear number of atypical
rounds while requiring the optimal memory shadow prices on the remaining
rounds to concentrate around a common value.

\section{Proof of Proposition~\ref{prop:independent-state-linear-regret}}
\label{app:independent-state-lower-bound}

Fix $B>0$ and $\alpha\in(0,1]$, and write
\[
    q:=\left\lfloor\frac{1}{\alpha}\right\rfloor,
    \qquad M:=q+1.
\]
Let $L,T>M$ be integers, and fix any possibly randomized,
non-anticipating allocator satisfying $a_t\in\mathcal A_t(B)$ at
every round. The fixed, non-anticipating refresh pipeline is the same
for the online allocator and the hindsight comparator. Partition the
$T$-round prefix into windows
of lengths $n_1,\ldots,n_W$, with $n_w=L$ for $w<W$,
$1\le n_W\le L$, and $T=\sum_{w=1}^{W}n_w$.
We construct a family of instances
$\{\mathcal I_{\boldsymbol z}:\boldsymbol z\in\{1,\ldots,M\}^{W}\}$,
where $z_w$ identifies the candidate that becomes relevant in window $w$.
We choose exogenous arrivals, so allocation decisions change memory states
but not the subsequent interaction sequence.
All raw candidate records and all auxiliary arrivals other than the
target-revelation signals are fixed independently of $\boldsymbol z$.
A window's target affects the interaction only through its revelation
and the subsequent utility profiles in that window.

\paragraph{A common prefix in each window.}
In the first $\min\{M,n_w\}$ rounds of window $w$, one new candidate
arrives per round. Candidates have distinct identities across windows,
and their raw contents are absent from earlier transcripts. Throughout
its arrival window, each candidate has raw size $\alpha B$; all
non-raw representations have
zero size and utility. This is a special case of the five-level
representation model and satisfies its monotonicity requirements.

All candidates have zero utility through local round $M$.
If $n_w>M$, the interaction reveals $z_w$ at round $M+1$ without
reproducing the candidate's raw contents. Its raw representation then
has utility one until the window ends, while all other representations
have utility zero. Carried history and items from earlier windows have
zero utility throughout the current window. Other arrivals have zero
utility and never reproduce a raw candidate. The oracle is exact, and
$0\le U_t(a)\le1$ for every feasible allocation on either policy's
memory trajectory.

The arrivals and oracle profiles in the first $\min\{M,n_w\}$ rounds
do not depend on $z_w$. Thus the allocator receives no information
about the target before its revelation. Allocation within a window
only selects representation levels for existing resident items; it
does not introduce new copies of discarded candidates. Together with
the irreversible representation constraint and the absence of
within-window archive access, this prevents recovery of a lost raw
representation before the window ends.
The allocator may carry its full internal state across boundaries;
no reset is imposed in the construction.

\paragraph{Feasibility and value of the hindsight benchmark.}
In each window with $n_w>M$, the hindsight policy deletes all carried
and non-target items and retains candidate $z_w$ in raw form from its
arrival until the window ends. In a window with $n_w\le M$, it deletes
all items. Its post-allocation occupancy satisfies
\[
    G_t^{\star}\le\alpha B\le B,
    \qquad t=1,\ldots,T,
    \qquad
    \sum_{t=1}^{T}g_t^{\star}\le\alpha T.
\]
This policy uses the same fixed refresh pipeline: the returned context
is deleted at the next allocation.

The policy earns $(n_w-M)_+$ in window $w$. No policy can earn more,
since each round after the first $M$ has utility at most one and all
earlier rounds have utility zero. Writing
$\operatorname{OPT}_T^{\mathrm{ind}}(\mathcal I_{\boldsymbol z})$
for the independent-state hindsight optimum, we obtain
\begin{equation}
\label{eq:app-independent-opt-value}
    \operatorname{OPT}_T^{\mathrm{ind}}(\mathcal I_{\boldsymbol z})
    =\sum_{w=1}^{W}(n_w-M)_+.
\end{equation}
This value holds for every realization of the boundary pipeline.

\paragraph{Online utility under fresh uncertainty.}
For the averaging argument, draw $Z_1,\ldots,Z_W$ independently and
uniformly from $\{1,\ldots,M\}$. Let $\xi$ be an exogenous random
tape driving the allocator and the fixed boundary pipeline, sampled
independently of $\boldsymbol Z=(Z_1,\ldots,Z_W)$. Conditional on
$\xi$, their responses are deterministic functions of the information
available to them.

Consider a window with $n_w>M$. Let $S_w$ be the set of its candidates
retained in raw form immediately after the local round-$M$ allocation.
Since $q\alpha B\le B<(q+1)\alpha B$, hard feasibility gives
$|S_w|\le q$. By non-anticipation and the common observable prefix,
$S_w$ is determined by $(\xi,Z_1,\ldots,Z_{w-1})$ and does not
depend on $Z_w$. Hence
\[
    \mathbb P\!\left(
        Z_w\in S_w\mid\xi,Z_1,\ldots,Z_{w-1}
    \right)
    =\frac{|S_w|}{M}
    \le\frac{q}{M}.
\]
The allocator earns positive utility after local round $M$ only if
$Z_w\in S_w$, and even then earns at most $n_w-M$. Therefore,
\[
    \mathbb E_{\boldsymbol Z,\xi}\!\left[
        \sum_{k=1}^{n_w}U_{w,k}(a_{w,k})
    \right]
    \le\frac{q}{M}(n_w-M)_+.
\]
The same inequality holds trivially when $n_w\le M$.
Summing over windows yields
\[
    \mathbb E_{\boldsymbol Z,\xi}\!\left[
        \sum_{t=1}^{T}U_t(a_t)
    \right]
    \le\frac{q}{M}\sum_{w=1}^{W}(n_w-M)_+.
\]

\paragraph{A fixed hard instance over the full interaction.}
For a fixed target vector $\boldsymbol z$, define
\[
    \rho(\boldsymbol z)
    :=\mathbb E_{\xi}\!\left[
        \operatorname{OPT}_T^{\mathrm{ind}}(\mathcal I_{\boldsymbol z})
        -\sum_{t=1}^{T}U_t(a_t)
    \right].
\]
Combining the preceding inequality with
\eqref{eq:app-independent-opt-value} gives
\[
    \frac{1}{M^W}
    \sum_{\boldsymbol z\in\{1,\ldots,M\}^{W}}\rho(\boldsymbol z)
    \ge
    \left(1-\frac{q}{M}\right)
    \sum_{w=1}^{W}(n_w-M)_+
    =\frac{1}{M}\sum_{w=1}^{W}(n_w-M)_+.
\]
Consequently, there exists a fixed vector $\boldsymbol z^{\star}$
such that, on $\mathcal I_{\boldsymbol z^{\star}}$,
\begin{equation}
\label{eq:app-window-dependent-independent-lower-bound}
    \operatorname{Reg}_T^{\mathrm{ind}}
    =\rho(\boldsymbol z^{\star})
    \ge\frac{1}{M}\sum_{w=1}^{W}(n_w-M)_+.
\end{equation}
The vector $\boldsymbol z^{\star}$ may depend on the allocator and
$T$, but not on the realized random seed.

\paragraph{A linear bound in the interaction length.}
Write $T=KL+r$, where $K=\lfloor T/L\rfloor$ and $0\le r<L$.
Then
\[
    \sum_{w=1}^{W}(n_w-M)_+
    =K(L-M)+(r-M)_+.
\]
Since $L\ge M+1$,
\[
    L-M\ge\frac{L}{M+1},
    \qquad
    (r-M)_+\ge\frac{r-M}{M+1}.
\]
Substituting into
\eqref{eq:app-window-dependent-independent-lower-bound} gives
\[
    \operatorname{Reg}_T^{\mathrm{ind}}
    \ge\frac{K(L-M)+(r-M)_+}{M}
    \ge\frac{KL+r-M}{M(M+1)}
    =\frac{T-M}{M(M+1)},
\]
which proves~\eqref{eq:independent-state-lower-bound}.
For fixed $\alpha$, $M$ is constant; in particular, for $T\ge2M$,
\[
    \operatorname{Reg}_T^{\mathrm{ind}}
    \ge\frac{T}{2M(M+1)}.
\]
Thus the worst-case expected regret is $\Omega_{\alpha}(T)$.

\section{Proof of Theorem~\ref{thm:global-guarantee}}
\label{app:global-performance-guarantee}

Fix an evaluation prefix $T$ and work on the probability-one event on which
Assumptions~\ref{ass:utility-oracle} and~\ref{ass:stability} hold up to $T$.
Set \(z_t:=g_t(a_t)-\alpha\).
Since $g_t(a_t)\in[0,1]$ and $\alpha\in(0,1]$, we have
$z_t\in[-\alpha,1-\alpha]\subseteq[-1,1]$.

\paragraph{Hard-capacity feasibility.}
Along MemPrice's realized trajectory,
$\mathcal A_t^{\mathrm{alg}}(B)=\mathcal A_t(B)$, and
Algorithm~\ref{alg:memprice} selects $a_t\in\mathcal A_t(B)$ at every round.
Hence
\[
    G_t(a_t)\le B,
    \qquad t=1,\ldots,T.
\]
Thus hard-capacity feasibility holds for every execution.

\paragraph{Global same-state comparison.}
Recall the priced frontier
\[
    \Phi_t(\lambda)
    :=
    \max_{a\in\mathcal A_t^{\mathrm{alg}}(B)}
    \left\{
        U_t(a)-\lambda\bigl(g_t(a)-\alpha\bigr)
    \right\}.
\]
For any $\lambda\ge0$ and any feasible sequence $(b_t)_{t=1}^T$ in the
global same-state benchmark,
\[
\begin{aligned}
    \sum_{t=1}^T U_t(b_t)
    &=
    \sum_{t=1}^T
    \left[
        U_t(b_t)-\lambda\bigl(g_t(b_t)-\alpha\bigr)
    \right]
    +
    \lambda\sum_{t=1}^T\bigl(g_t(b_t)-\alpha\bigr)\\
    &\le
    \sum_{t=1}^T \Phi_t(\lambda),
\end{aligned}
\]
where the inequality uses
$\sum_{t=1}^T g_t(b_t)\le\alpha T$.
For the fixed prefix $T$, let $\lambda_T^\star$ be a common price satisfying Assumption~\ref{ass:stability}, and it holds that
\begin{equation}
\label{eq:app-global-dual-bound}
    \operatorname{OPT}_T^{\mathrm{gss}}
    \le
    \sum_{t=1}^T \Phi_t(\lambda^\star).
\end{equation}

At round $t$, exact maximization of the estimated priced objective and
Assumption~\ref{ass:utility-oracle} imply
\begin{equation}
\label{eq:app-frontier-comparison}
   U_t(a_t)
-
\lambda_t\bigl(g_t(a_t)-\alpha\bigr)
\ge
\Phi_t(\lambda_t)-2e_t.
\end{equation}
Indeed, the additive term $\lambda_t\alpha$ is independent of the action, and
the definition of $e_t$ implies that replacing estimated utilities by true
utilities loses at most $2e_t$.

Combining~\eqref{eq:app-global-dual-bound} and
\eqref{eq:app-frontier-comparison} gives, pathwise,
\begin{align}
    \operatorname{OPT}_T^{\mathrm{gss}}
    -
    \sum_{t=1}^T U_t(a_t)
    &\le
    \sum_{t=1}^T
    \bigl[\Phi_t(\lambda^\star)-\Phi_t(\lambda_t)\bigr]
    -
    \sum_{t=1}^T \lambda_t z_t
    +
    2\sum_{t=1}^T e_t
    \notag\\
    &\le
    \Delta_T
    -
    \sum_{t=1}^T \lambda_t z_t
    +
    2\sum_{t=1}^T e_t .
\label{eq:app-global-pathwise-comparison}
\end{align}
where the second inequality follows from
$\Phi_t(\lambda_t)\ge\inf\limits_{\lambda\ge0}\Phi_t(\lambda)$ and
Assumption~\ref{ass:stability}.

\paragraph{Global queue drift.}
The queue recursion gives
\[
    Q_{t+1}^2
    =
    [Q_t+z_t]_+^2
    \le
    Q_t^2+2Q_tz_t+z_t^2.
\]
Using $\lambda_t=\eta_tQ_t$ and $z_t^2\le1$,
\[
    -\lambda_tz_t
    \le
    \frac{\eta_t}{2}
    \bigl(Q_t^2-Q_{t+1}^2\bigr)
    +
    \frac{\eta_t}{2}.
\]
Since $\eta_t=c/\sqrt t$ is nonincreasing and $Q_1=0$,
\[
\begin{aligned}
    \sum_{t=1}^T
    \eta_t\bigl(Q_t^2-Q_{t+1}^2\bigr)
    &=
    \eta_1Q_1^2
    +
    \sum_{t=2}^T
    (\eta_t-\eta_{t-1})Q_t^2
    -
    \eta_TQ_{T+1}^2\\
    &\le0.
\end{aligned}
\]
Therefore,
\begin{equation}
\label{eq:app-global-drift-bound}
    -\sum_{t=1}^T\lambda_tz_t
    \le
    \frac12\sum_{t=1}^T\eta_t
    =
    \frac c2\sum_{t=1}^T\frac1{\sqrt t}
    \le
    c\sqrt T.
\end{equation}
Combining~\eqref{eq:app-global-pathwise-comparison} and
\eqref{eq:app-global-drift-bound} yields
\[
    \operatorname{OPT}_T^{\mathrm{gss}}
    -
    \sum_{t=1}^T U_t(a_t)
    \le
    c\sqrt T
    +
    2\sum_{t=1}^T e_t
    +
    \Delta_T.
\]
Taking expectations and using $\mathcal E_T=\sum_{t=1}^T \mathbb E[e_t]$ gives
\begin{equation}
\label{eq:app-global-regret-bound}
    \operatorname{Reg}_T^{\mathrm{gss}}
    \le
    c\sqrt T
    +
    2\mathcal E_T
    +
    \Delta_T.
\end{equation}

\paragraph{Controlling occupancy through the all-delete action.}
Let $a_t^0$ denote the all-delete allocation. It is feasible and satisfies
\[
    U_t(a_t^0)=0,
    \qquad
    g_t(a_t^0)=0.
\]
Exact maximization of the estimated priced objective gives
\[
    \widehat U_t(a_t)-\lambda_tg_t(a_t)
    \ge
    \widehat U_t(a_t^0).
\]
Assumption~\ref{ass:utility-oracle} therefore implies
\[
\begin{aligned}
   \lambda_tg_t(a_t)
&\le
\widehat U_t(a_t)-\widehat U_t(a_t^0)\\
&\le
U_t(a_t)-U_t(a_t^0)+2e_t\\
&\le
\overline U+2\overline\epsilon
\qquad\text{a.s.}
\end{aligned}
\]
Set
\[
    C_U:=\overline U+2\overline\epsilon,
    \qquad
    \Theta_t
    :=
    \frac{C_U}{\alpha\eta_t}
    =
    \frac{C_U}{\alpha c}\sqrt t.
\]
The sequence $\Theta_t$ is nondecreasing. Whenever $Q_t>\Theta_t$,
\[
    \lambda_t
    =
    \eta_tQ_t
    >
    \frac{C_U}{\alpha},
\]
and hence
\[
    g_t(a_t)
    \le
    \frac{C_U}{\lambda_t}
    <
    \alpha.
\]
Thus $Q_{t+1}\le Q_t$ whenever $Q_t>\Theta_t$.

We claim that
\begin{equation}
\label{eq:app-queue-threshold-bound}
    Q_{t+1}\le\Theta_t+1,
    \qquad t\ge1.
\end{equation}
For $t=1$,
\[
    Q_2
    =
    [g_1(a_1)-\alpha]_+
    \le1
    \le\Theta_1+1.
\]
Suppose $t\ge2$ and $Q_t\le\Theta_{t-1}+1$.
If $Q_t\le\Theta_t$, then
\[
    Q_{t+1}
    \le
    Q_t+1
    \le
    \Theta_t+1.
\]
If $Q_t>\Theta_t$, then
\[
    Q_{t+1}
    \le
    Q_t
    \le
    \Theta_{t-1}+1
    \le
    \Theta_t+1.
\]
This proves~\eqref{eq:app-queue-threshold-bound}, and in particular
\begin{equation}
\label{eq:app-global-queue-bound}
    Q_{T+1}
    \le
    \frac{\overline U+2\overline\epsilon}{\alpha c}\sqrt T
    +1.
\end{equation}

\paragraph{Cumulative occupancy violation.}
Since $[x]_+\ge x$,
\[
    Q_{t+1}\ge Q_t+z_t.
\]
Summing from $t=1$ to $T$ and using $Q_1=0$ gives
\[
    Q_{T+1}
    \ge
    \sum_{t=1}^T z_t.
\]
Since $Q_{T+1}\ge0$,
\[
    \left[\sum_{t=1}^Tz_t\right]_+
    \le
    Q_{T+1}.
\]
Therefore, by~\eqref{eq:app-global-queue-bound},
\[
\begin{aligned}
    \operatorname{Viol}_T
    &=
    B\left[\sum_{t=1}^Tz_t\right]_+\\
    &\le
    BQ_{T+1}\\
    &\le
    B\left(
        \frac{\overline U+2\overline\epsilon}{\alpha c}\sqrt T
        +1
    \right)
\end{aligned}
\]
almost surely. This proves all claims of
Theorem~\ref{thm:global-guarantee}.

\section{Proof of Theorem~\ref{thm:expected-global-guarantee}}
\label{app:expected_proof}
Fix an evaluation prefix \(T\), and write
\[
    z_t:=g_t(a_t)-\alpha .
\]

\paragraph{Expectation-feasible hindsight comparison.}
Let \(\pi\in\Pi_T^{\mathrm{egss}}\) be any expectation-feasible
same-state hindsight selector. By definition,
\[
    b_t^\pi(\omega,\zeta)
    \in
    \mathcal A_t^{\mathrm{alg}}(B;\omega)
\]
for every round \(t\) almost surely, and
\[
    \mathbb E\!\left[
        \sum_{t=1}^T g_t(b_t^\pi)
    \right]
    \le
    \alpha T .
\]
For the deterministic price \(\lambda_T^\star\) in
Assumption~\ref{ass:expected-stability},
\[
\begin{aligned}
    \mathbb E\!\left[
        \sum_{t=1}^T U_t(b_t^\pi)
    \right]
    &=
    \mathbb E\!\left[
        \sum_{t=1}^T
        \left(
            U_t(b_t^\pi)
            -
            \lambda_T^\star
            \bigl(g_t(b_t^\pi)-\alpha\bigr)
        \right)
    \right] \\
    &\quad+
    \lambda_T^\star
    \mathbb E\!\left[
        \sum_{t=1}^T
        \bigl(g_t(b_t^\pi)-\alpha\bigr)
    \right] \\
    &\le
    \mathbb E\!\left[
        \sum_{t=1}^T
        \Phi_t(\lambda_T^\star)
    \right].
\end{aligned}
\]
The last inequality uses
\(\lambda_T^\star\ge0\), expectation feasibility, and
\(b_t^\pi\in\mathcal A_t^{\mathrm{alg}}(B)\) almost surely.
Taking the supremum over
\(\pi\in\Pi_T^{\mathrm{egss}}\) therefore gives
\begin{equation}
\label{eq:app-expected-dual}
    \operatorname{OPT}_T^{\mathrm{egss}}
    \le
    \mathbb E\!\left[
        \sum_{t=1}^T
        \Phi_t(\lambda_T^\star)
    \right].
\end{equation}

\paragraph{Online frontier comparison.}
At each round, MemPrice exactly maximizes
\[
    \widehat U_t(a)-\lambda_t g_t(a)
\]
over \(\mathcal A_t^{\mathrm{alg}}(B)\).
Under Assumption~\ref{ass:utility-oracle}, the same argument as in the
proof of Theorem~\ref{thm:global-guarantee} gives, almost surely,
\begin{equation}
\label{eq:app-expected-frontier}
   U_t(a_t)
-
\lambda_t\bigl(g_t(a_t)-\alpha\bigr)
\ge
\Phi_t(\lambda_t)-2e_t
\qquad\text{a.s.}
\end{equation}
Equivalently,
\[
    U_t(a_t)
\ge
\Phi_t(\lambda_t)
+
\lambda_t z_t
-
2e_t.
\]
Summing over \(t\), taking expectations, and using
\(\mathcal E_T=\sum_{t=1}^T\mathbb E[e_t]\) yields
\begin{equation}
\label{eq:app-expected-online}
    \mathbb E\!\left[
        \sum_{t=1}^T U_t(a_t)
    \right]
    \ge
    \mathbb E\!\left[
        \sum_{t=1}^T \Phi_t(\lambda_t)
    \right]
    +
    \mathbb E\!\left[
        \sum_{t=1}^T\lambda_t z_t
    \right]
    -
    2\mathcal E_T .
\end{equation}

Combining
Eqs.~\eqref{eq:app-expected-dual}
and~\eqref{eq:app-expected-online},
\begin{align}
    \operatorname{Reg}_T^{\mathrm{egss}}
    &\le
    \mathbb E\!\left[
        \sum_{t=1}^T
        \bigl(
            \Phi_t(\lambda_T^\star)
            -
            \Phi_t(\lambda_t)
        \bigr)
    \right]
    -
    \mathbb E\!\left[
        \sum_{t=1}^T\lambda_t z_t
    \right]
    +
    2\mathcal E_T .
\label{eq:app-expected-regret-decomp}
\end{align}
Since
\[
    \Phi_t(\lambda_t)
    \ge
    \inf_{\lambda\ge0}\Phi_t(\lambda),
\]
Assumption~\ref{ass:expected-stability} implies
\begin{equation}
\label{eq:app-expected-frontier-deviation}
    \mathbb E\!\left[
        \sum_{t=1}^T
        \bigl(
            \Phi_t(\lambda_T^\star)
            -
            \Phi_t(\lambda_t)
        \bigr)
    \right]
    \le
    \Delta_T^{\mathrm E}.
\end{equation}

\paragraph{Queue drift.}
The queue recursion
\[
    Q_{t+1}=[Q_t+z_t]_+
\]
implies, pathwise,
\[
    Q_{t+1}^2
    \le
    Q_t^2+2Q_tz_t+z_t^2 .
\]
Since \(z_t^2\le1\),
\(\lambda_t=\eta_tQ_t\), and
\(\eta_t=c/\sqrt t\),
\[
    -\lambda_tz_t
    \le
    \frac{\eta_t}{2}
    \bigl(Q_t^2-Q_{t+1}^2\bigr)
    +
    \frac{\eta_t}{2}.
\]
Because \(\eta_t\) is nonincreasing and \(Q_1=0\),
\[
    \sum_{t=1}^T
    \eta_t
    \bigl(Q_t^2-Q_{t+1}^2\bigr)
    \le0.
\]
Hence, almost surely,
\[
    -\sum_{t=1}^T\lambda_tz_t
    \le
    \frac12\sum_{t=1}^T\eta_t
    \le
    c\sqrt T .
\]
Taking expectations preserves the bound:
\begin{equation}
\label{eq:app-expected-drift}
    -
    \mathbb E\!\left[
        \sum_{t=1}^T\lambda_tz_t
    \right]
    \le
    c\sqrt T .
\end{equation}

Combining
Eqs.~\eqref{eq:app-expected-regret-decomp},
\eqref{eq:app-expected-frontier-deviation}, and
\eqref{eq:app-expected-drift} gives
\[
    \operatorname{Reg}_T^{\mathrm{egss}}
    \le
    c\sqrt T
    +
    2\mathcal E_T
    +
    \Delta_T^{\mathrm E},
\]
which proves Eq.~\eqref{eq:expected-global-regret}.

\paragraph{Expected average occupancy.}
It remains to establish
Eq.~\eqref{eq:expected-average-occupancy}.
As in the proof of Theorem~\ref{thm:global-guarantee}, comparison with the
all-delete allocation and Assumption~\ref{ass:utility-oracle} give
\[
    \lambda_tg_t(a_t)
    \le
    \overline U+2\overline\epsilon .
\]
Let
\[
    C_U:=\overline U+2\overline\epsilon,
    \qquad
    \Theta_t
    :=
    \frac{C_U}{\alpha\eta_t}
    =
    \frac{C_U}{\alpha c}\sqrt t .
\]
Whenever \(Q_t>\Theta_t\),
\[
    \lambda_t
    =
    \eta_tQ_t
    >
    \frac{C_U}{\alpha},
\]
and therefore
\[
    g_t(a_t)
    \le
    \frac{C_U}{\lambda_t}
    <
    \alpha .
\]
Thus \(Q_{t+1}\le Q_t\) whenever \(Q_t>\Theta_t\).
The same induction as in
Theorem~\ref{thm:global-guarantee} gives
\[
    Q_{T+1}
    \le
    \frac{\overline U+2\overline\epsilon}
         {\alpha c}
    \sqrt T
    +1
    \qquad\text{a.s.}
\]
Moreover, since
\[
    Q_{t+1}
    =
    [Q_t+z_t]_+
    \ge
    Q_t+z_t,
\]
summing over \(t\) gives
\[
    \sum_{t=1}^T
    \bigl(g_t(a_t)-\alpha\bigr)
    =
    \sum_{t=1}^T z_t
    \le
    Q_{T+1}.
\]
Hence, almost surely,
\[
    \frac1T\sum_{t=1}^T g_t(a_t)
    \le
    \alpha
    +
    \frac{\overline U+2\overline\epsilon}
         {\alpha c\sqrt T}
    +
    \frac1T .
\]
Taking expectations yields
\[
    \mathbb E\!\left[
        \frac1T\sum_{t=1}^T g_t(a_t)
    \right]
    \le
    \alpha
    +
    \frac{\overline U+2\overline\epsilon}
         {\alpha c\sqrt T}
    +
    \frac1T ,
\]
which proves Eq.~\eqref{eq:expected-average-occupancy}.

\section{Implementation Details}
\subsection{Detailed Experimental Settings}
\label{app:experimental-settings}

\paragraph{Information regimes.}
Our experiments distinguish two memory-allocation regimes according to when
the downstream information need becomes observable.
In \textbf{Setting I}, memory is constructed without access to the terminal
query. For a chronological history \(H\), the allocator produces a bounded
memory state before the target question is released. Target questions,
answers, evidence annotations, question types, and evaluation feedback are
therefore unavailable to both the allocator and its utility estimator during
memory construction. This setting evaluates whether a memory policy can
preserve information that is useful for unknown future requests.
In \textbf{Setting II}, the task goal is available from reset and may be used
to determine which parts of the interaction history should be retained.
Future observations, tool outputs, environment states, and evaluator-only
information remain unavailable. This setting corresponds to long-horizon
agents that know what they are trying to accomplish but must manage an
increasing trajectory online.

\paragraph{LongMemEval-S.}
LongMemEval~\citep{wu2025longmemeval} evaluates long-term interactive memory
over histories containing multiple temporally separated sessions.
We use the \textbf{LongMemEval-S} variant as a primary query-blind benchmark.
For each history, the complete chronological interaction stream is processed
before its associated evaluation questions are released.
All questions attached to the same history therefore reuse the same frozen
terminal memory state rather than reconstructing memory separately for each
question.
This prevents the allocator from specializing memory to individual queries.
We report the benchmark's official QA accuracy as the primary metric.
The larger LongMemEval-M variant is reserved for long-context stress analysis
rather than the primary comparison.

\paragraph{LoCoMo.}
LoCoMo~\citep{maharana2024locomo} evaluates very long-term conversational
memory across extended multi-session dialogues.
Its questions cover several forms of retrieval and reasoning, including
single-hop factual recall, temporal reasoning, multi-hop questions, and cases
requiring information to be integrated across distant portions of a
conversation.
We expose only the chronological conversation while memory is constructed;
evaluation questions and their annotations are revealed after the memory state
has been frozen.
The complete conversation is treated as the evaluation unit, and all methods
receive identical session segmentation and memory budgets.
We use normalized partial-match F1 as the primary evaluation metric.

\paragraph{AppWorld.}
AppWorld~\citep{trivedi2024appworld} is a stateful interactive benchmark in
which an agent operates a collection of applications through structured APIs
to satisfy a natural-language goal.
Unlike Setting-I benchmarks, the goal is visible before the first action, so
memory policies may condition retention decisions on task relevance.
The agent does not observe future tool results, future application state, or
the benchmark evaluator.
Each completed action together with its resulting observation forms a
historical interaction unit for memory management.
We evaluate separately on the official Test-Normal and Test-Challenge splits.
The primary metric is \textbf{Task Goal Completion (TGC)}, for which a task is
successful only when its benchmark-native state-based tests are satisfied.
Scenario Goal Completion and difficulty-based breakdowns are retained as
secondary analyses.

\paragraph{WebShop.}
WebShop~\citep{yao2022webshop} provides a reproducible simulated e-commerce
environment in which an agent must search, inspect, and select products
satisfying a user-specified shopping goal.
The complete goal is available from reset, while future search results,
product pages, and environment transitions are revealed only through the
agent's own actions.
WebShop therefore provides a second goal-observable setting with a different
interaction structure from AppWorld.
Historical search and browsing observations are subject to the same bounded
memory interface as other Setting-II tasks.
We report \textbf{Success Rate}, the fraction of tasks receiving the maximum
terminal reward, as the primary metric and average terminal reward as a
secondary metric.

\paragraph{Setting-I baselines.}
All Setting-I competitors obey the same query-blind information contract and
hard historical-memory budget.
\textbf{Recent/FIFO} retains history in reverse chronological order until the
budget is exhausted and provides a pure recency control
(cf.~\citealp{park2023generative}).
\textbf{Uniform Allocation} uses the same candidate representation levels as
our method but distributes resources uniformly across memory items, isolating
the effect of adaptive allocation.
\textbf{Content Salience} ranks memory using query-independent lexical
salience, while \textbf{Dense Salience} uses a dense semantic representation;
both retain the highest-valued content under the common capacity.
\textbf{LRE-SelfSup}~\citep{lia2026lre} learns to predict the future usefulness
of history units without observing the future query.
\textbf{Full Context} is reported separately as an unbudgeted upper-reference
condition and is not included in matched-resource comparisons.

\paragraph{Setting-II baselines.}
\textbf{ReAct/FIFO History}~\citep{yao2023react} uses the standard iterative
reasoning-and-action interface while retaining the most recent complete
trajectory items under budget.
\textbf{Summary Agent} periodically compresses accumulated trajectory history
into a global summary, providing a conventional long-horizon context
compression control.
\textbf{ZipAct-B} is our resource-matched adaptation of
ZipAct~\citep{pan2026zipact}.
It preserves ZipAct's recurrent Goal--World--Constraint state update but
requires the state visible to the actor to satisfy the same
tokenizer-measured hard historical-memory budget \(B\) before every action.
Thus, differences in task performance cannot be attributed simply to one
method receiving a larger actor-visible context.

\paragraph{Historical-memory budget.}
The resource budget \(B\) applies only to historical information made visible
to the reader or agent at a decision point.
The current query or task goal, current observation, system instructions, tool
schemas, and reserved output tokens are handled separately when enforcing the
physical context-window constraint.
Dataset-specific budget grids are determined without using test outcomes from
development-history length statistics and available physical-context
headroom.
We use four budget levels derived from the corresponding maximum feasible
budget \(B_{\max}\), and report the primary single-budget comparison at
\(B^\star=B_{\max}/2\).
All methods within a dataset use exactly the same budget values.

\paragraph{Multi-fidelity allocation.}
Each memory item is represented by five ordered fidelity levels:
\emph{raw}, \emph{summary}, \emph{abstraction}, \emph{keywords}, and
\emph{delete}.
Candidate representations are generated only from the currently available
text and cannot restore information already removed from an active
representation.
Successive compressed representations target \(0.6\) of their parent length.
Given candidate utilities and tokenizer-measured sizes, our method jointly
selects one representation per item through an exact multiple-choice knapsack
problem subject to the hard capacity \(B\).

The utility estimator obeys the information contract of its setting.
For Setting I it evaluates general-purpose future retention utility without
access to the terminal query, whereas in Setting II it may condition utility
on the task goal already visible to the agent.
Utility scores are constrained to \([0,1]\), delete has zero utility, and
candidate utilities are monotonically repaired across decreasing fidelity
levels when necessary.

\paragraph{Adaptive pricing and memory maintenance.}
We target average normalized occupancy \(\alpha=0.8\) and maintain a persistent
pressure state
\[
Q_{t+1}
=
\left[
Q_t+\frac{G_t}{B}-\alpha
\right]_+,
\]
where \(G_t\) denotes the selected historical-memory size.
The corresponding adaptive price is
\[
\lambda_t=\frac{cQ_t}{\sqrt{t}}.
\]
Both \(Q_t\) and the global time index persist across memory-maintenance
boundaries.
The scale \(c\) is chosen separately for the two information regimes using
development data only, minimizing the deviation between mean occupancy and
the target \(\alpha\); downstream task accuracy is never used for this
calibration.

Allocation is updated as new memory items are admitted.
A maintenance boundary occurs when either active memory reaches capacity or
the number of interactions since the previous boundary reaches \(L=10\).
At a boundary, complete raw history is first copied to an append-only external
archive, the active memory is compressed, and relevant archived chunks may
then be recovered into the next active context.
The archive is persistent storage and does not itself count toward the
reader-visible budget \(B\); any retrieved text does.
Recovery uses BM25 with \(k_1=1.5\), \(b=0.75\), and top-\(8\) retrieval,
subject again to the hard active-memory capacity.

\paragraph{Replication.}
Every main configuration is evaluated with three shared random seeds, namely 2026, 2027, and 2028.
The same seed set is used for all competing policies on a given benchmark so that stochastic variation is paired across methods.
Language-model decoding itself remains deterministic where supported.
We report the mean across the three runs together with dispersion in the corresponding result tables.

\end{document}